\documentclass[final]{cvpr}

\usepackage{epsfig}
\usepackage{enumitem}
\usepackage{amsmath,amsfonts,amsthm,amssymb}
\usepackage{xspace}
\usepackage[utf8]{inputenc} 
\usepackage[T1]{fontenc}    
\usepackage{url}            
\usepackage{booktabs}       
\usepackage{amsfonts}       
\usepackage{nicefrac}       
\usepackage{microtype}      
\usepackage{multirow}
\usepackage{times}
\usepackage{enumerate}
\usepackage{lscape,color,url}
\usepackage{adjustbox}
\usepackage{graphicx,graphics,wrapfig,epstopdf,subfig}
\usepackage{algorithm}
\usepackage{algorithmic}
\usepackage{overpic}
\usepackage{mathtools}
\usepackage{soul}
\usepackage{caption}

\usepackage{amsmath,amsfonts,bm}

\def\eqref#1{equation~\ref{#1}}

\def\1{\bm{1}}

\def\ervu{{\textnormal{u}}}
\def\ervv{{\textnormal{v}}}

\def\vv{{\bm{v}}}

\DeclareMathAlphabet{\mathsfit}{\encodingdefault}{\sfdefault}{m}{sl}
\SetMathAlphabet{\mathsfit}{bold}{\encodingdefault}{\sfdefault}{bx}{n}

\newcommand{\Amat}{{\bf A}}

\newcommand{\Cmat}{{\bf C}}

\newcommand{\Qmat}{{\bf Q}}

\newcommand{\hv}{{\boldsymbol h}}

\newcommand{\pv}{{\boldsymbol p}}
\newcommand{\qv}{{\boldsymbol q}}

\newcommand{\uv}{{\boldsymbol u}}

\newcommand{\xv}{{\boldsymbol x}}
\newcommand{\yv}{{\boldsymbol y}}

\newcommand{\zv}{{\boldsymbol z}}

\newcommand{\betav}{{\boldsymbol \beta}}

\newcommand{\thetav}{{\boldsymbol \theta}}

\newcommand{\muv}{{\boldsymbol \mu}}
\newcommand{\nuv}{{\boldsymbol \nu}}

\newcommand{\piv}{{\boldsymbol \pi}}

\newcommand{\phiv}{{\boldsymbol \phi}}

\newcommand{\Xcal}{\mathcal{X}}
\newcommand{\Ycal}{\mathcal{Y}}

\newcommand{\Lcal}{\mathcal{L}}

\newcommand{\Bcal}{\mathcal{B}}

\newcommand{\Wcal}{\mathcal{W}}

\theoremstyle{definition}
\newtheorem{definition}{Definition}[section]
 
\newcommand*{\thead}[1]{%
\multicolumn{1}{c}{\bfseries\begin{tabular}{@{}c@{}}#1\end{tabular}}}

\newcommand*{\mathcolor}{}
\def\mathcolor#1#{\mathcoloraux{#1}}
\newcommand*{\mathcoloraux}[3]{%
  \protect\leavevmode
  \begingroup
    \color#1{#2}#3%
  \endgroup
}

\newcommand{\Ep}{{\mathbb E}}

\usepackage[pagebackref=true,breaklinks=true,colorlinks,bookmarks=false]{hyperref}

\def\cvprPaperID{10552} 
\def\confYear{CVPR 2021}

\begin{document}

\title{Wasserstein Contrastive Representation Distillation}

\author{Liqun Chen$^{1}$\thanks{Equal contribution}, Dong Wang$^{1*}$, Zhe Gan$^2$,  Jingjing Liu$^2$, Ricardo Henao$^1$, Lawrence Carin$^1$\\
$^1$Duke University, \quad $^2$Microsoft Corporation\\
{\tt\small \{liqun.chen, dong.wang, ricardo.henao, lcarin\}@duke.edu, \{zhe.gan, jingjl\}@microsoft.com}
}

\maketitle

\begin{abstract}
The primary goal of knowledge distillation (KD) is to encapsulate the information of a model learned from a teacher network into a student network, with the latter being more compact than the former. Existing work, e.g., using Kullback-Leibler divergence for distillation, may fail to capture important structural knowledge in the teacher network and often lacks the ability for feature generalization, particularly in situations when teacher and student are built to address different classification tasks. We propose Wasserstein Contrastive Representation Distillation (WCoRD), which leverages both primal and dual forms of Wasserstein distance for KD. The dual form is used for global knowledge transfer, yielding a contrastive learning objective that maximizes the lower bound of mutual information between the teacher and the student networks. The primal form is used for local contrastive knowledge transfer within a mini-batch, effectively matching the distributions of features between the teacher and the student networks. Experiments demonstrate that the proposed WCoRD method outperforms state-of-the-art approaches on privileged information distillation, model compression and cross-modal transfer.
\end{abstract}
\vspace{-5mm}
\section{Introduction}

The recent success of deep learning methods has brought about myriad efforts to apply them beyond benchmark datasets, but a number of challenges can emerge in real-world scenarios.
For one, as the scale of deep learning models continues to grow (\textit{e.g.}, \cite{he2016identity, devlin2018bert}), it has become increasingly difficult to deploy such trained networks on more computationally-restrictive platforms, such as smart phones, remote sensors, and edge devices.
Additionally, deep networks require abundant data for training, but large datasets are often private \cite{ribli2018detecting}, classified \cite{liang2018automatic}, or institutional \cite{sun2017revisiting}, which the custodians may be hesitant to release publicly.
Labeled datasets in specialized domains may also be rare or expensive to produce. Finally, despite ample datasets from neighboring modalities, conventional frameworks lack clear ways to leverage cross-modal data.

Knowledge distillation (KD), which has become an increasingly important topic in the deep learning community, offers a potential solution to these challenges.
In KD, the goal is to improve a \textit{student} model's performance by supplementing it with additional feedback from a \textit{teacher} model.
Often the teacher has larger capacity than the student, or has access to additional data that the student does not.
As such, KD can transfer this additional and valuable knowledge from the teacher to the student.
In early KD methods \cite{hinton2015distilling}, this supplemental supervision was imposed by asking the student to minimize the Kullback-Leibler (KL) divergence between its output prediction distribution and the teacher's. Given that the prediction probability distribution contains richer and more informative signals than the one-hot labels, student models have been shown to benefit from this extra supervision.
However, the low dimensionality of prediction distribution means that the amount of information encoded (therefore transferred) can be limited. For cross-modal transfer, these predictions may even be irrelevant, making KL divergence unable to transfer meaningful information.

In contrast, intermediate representations present an opportunity for more informative learning signals, as a number of recent works have explored \cite{romero2014fitnets, zagoruyko2016paying,sun2019patient,tian2019contrastive,sun2020contrastive}.
However, as observed by \cite{tian2019contrastive}, these methods often compare poorly with the basic KD, potentially due to the challenge of defining a proper distance metric between features of the teacher and student networks. 
Furthermore, they heavily rely on strategies to copy teacher's behavior, \emph{i.e.}, aligning the student's outputs to those from the teacher. We argue that such practice overlooks a key factor: the teacher's experience may not necessarily generalize well to the student. 

Motivated by this, we present \textbf{W}asserstein \textbf{Co}nstrastive \textbf{R}epresentation \textbf{D}istillation (WCoRD),  a new KD framework that reduces the generalization gap between teacher and student to approach better knowledge transfer. 
Specifically, our approach constitutes \emph{distillation} and \emph{generalization} blocks, realized by solving the \emph{dual} and \emph{primal} form of the Wasserstein distance (WD), respectively.
For better distillation, we leverage the dual form of WD to maximize the mutual information (MI) between student and teacher representation distributions, using an objective inspired by Noise Contrastive Estimation (NCE) \cite{gutmann2010noise}. 
Unlike previous methods~\cite{tian2019contrastive}, we propose to impose a 1-Lipschitz constraint to the critic via spectral normalization~\cite{miyato2018spectral}.
By shifting the critic to one based on optimal transport, we improve stability and sidestep some of the pitfalls of KL divergence minimization \cite{chen2017symmetric, mcallester2020formal}.
We term this as \textit{global} contrastive knowledge transfer.

For better generalization, we also use the primal form of WD to indirectly bound generalization error via regularizing the Wasserstein distance between the feature distributions of the student and teacher. 
This results in a relaxation that allows for coupling student and teacher features across multiple examples within each mini-batch, as opposed to the one-to-one matching in previous methods (\emph{i.e.}, strictly copying the teacher's behavior). In principle, this serves to directly match the feature distributions of the student and teacher networks. We term this \emph{local} contrastive knowledge transfer.
With the use of both primal and dual forms, we are able to maximize MI and simultaneously minimize the feature distribution discrepancy.

The main contributions are summarized as follows. ($i$) We present a novel Wasserstein learning framework for representation distillation, utilizing the dual and primal forms of the Wasserstein distance for \emph{global} constrastive learning and \emph{local} feature distribution matching, respectively. 
($ii$) To demonstrate the superiority of the proposed approach, we first conduct comprehensive experiments on benchmark datasets for model compression and cross-modal transfer. To demonstrate versatility, we further apply our method to a real-world dataset for privileged information distillation.

\section{Background}
\subsection{Knowledge Distillation}\label{sec:model}
In knowledge distillation, a student network is trained by leveraging additional supervision from a trained teacher network. Given an input sample $(\bm x, y)$, where $\bm x$ is the network input and $y$ is the one-hot label, the distillation objective encourages the output probability distribution over predictions from the student and teacher networks to be similar. Assume $\zv^T$ and $\zv^S$ are the logit representations (before the softmax layer) of the teacher and student network, respectively. In standard KD~\cite{hinton2015distilling}, the training of the student network involves two supervised loss terms:
\begin{align}\label{eq:distillation_general_loss}
    \mathcal{L} = & \, \mathcal{L}_{\mathrm{CE}}\left(y, \mathrm{softmax}(\zv^S) \right) \nonumber \\&+ \alpha \cdot \mathrm{KL}\left( \mathrm{softmax}(\zv^T / \rho) \| \mathrm{softmax}(\zv^S / \rho)  \right)\,, 
\end{align}
where $\rho$ is the temperature, and $\alpha$ is the balancing weight.
The representation in (\ref{eq:distillation_general_loss}) is optimized with respect to the student network parameters $\bm \theta_S$, while the teacher network (parameterized by $\bm \theta_T$) is pre-trained and fixed. The first term in (\ref{eq:distillation_general_loss}) enforces label supervision, which is conventionally implemented as a cross-entropy loss for classification tasks. The second term encourages the student network to produce distributionally-similar outputs to the teacher network. However, there are inevitable limitations to this approach. Deep neural networks learn structured features through the intermediate hidden layers, capturing spatial or temporal correlations of the input data.
These representations are then collapsed to a low-dimensional prediction distribution, losing this complex structure.
Furthermore, 
the KL divergence used here can be unstable numerically due to its asymmetry~\cite{chen2017symmetric, ozair2019wasserstein}. For example, it can overly concentrate on small details: a small imperfection can put a sample outside the distribution and explode the KL toward infinity.
Despite this, KD objectives based on KL divergence can still be effective and remain popular.





We aim to provide a general and principled framework for distillation based on \emph{Wasserstein distance}, 
where both global contrastive learning and local distribution matching are introduced to facilitate  knowledge transfer to the student.
By using the Wasserstein metric, we also avoid some of the drawbacks of KL-based approaches.
Note that our approach utilizes feature representations at the penultimate layer (before logits), denoted as $\hv^S$ and $\hv^T$ for the student and teacher networks, respectively.

\subsection{Wasserstein Distance}\label{subsec:wasserstein_intro}


One of the more recently-proposed distance measures in knowledge distillation is the contrastive loss~\cite{tian2019contrastive}. The goal is to move similar samples closer while pushing different ones apart in the feature space (\emph{i.e.}, $\zv^S$ and $\zv^T$, or $\hv^S$ and $\hv^T$). 
We further extend and generalize the idea of contrastive loss with Wasserstein Distance (a.k.a. Earth Mover's Distance, or Optimal Transport Distance).
In the following, we give a brief introduction to the \emph{primal} and \emph{dual} forms of the general Wasserstein Distance (WD). 
The primal form~\cite{villani2008optimal} is defined as follows.
\begin{definition} \label{def:wd}
    Consider two probability distribution: $\xv_1\sim \pv_1$, and $\xv_2\sim \pv_2$.
    The Wasserstein-1 distance between $\pv_1$ and $\pv_2$ can be formulated as:
    \begin{align}\label{eq:wasserstein_primal}
        \Wcal(\pv_1, \pv_2) = \inf_{\pi \in \Pi(\pv_1, \pv_2)} \int_{\mathcal{M} \times \mathcal{M}} c( \xv_1, \xv_2) d\pi(\xv_1, \xv_2) \,, \nonumber
    \end{align}
    where $c(\cdot)$ is a point-wise cost function evaluating the distance between $\xv_1$ and $\xv_2$, and $\Pi(\pv_1, \pv_2)$ is the set of all possible couplings of $\pv_1(\xv_1)$ and $\pv_2(\xv_2)$; $\mathcal{M}$ is the space of $\xv_1$ and $\xv_2$, and $\pi(\xv_1, \xv_2)$ is a joint distribution satisfying $\int_{\mathcal{M}}\pi(\xv_1, \xv_2)d \bm \xv_2 = \pv_1(\xv_1)$ and $\int_{\mathcal{M}}\pi(\xv_1, \xv_2)d \xv_1 = \pv_2(\xv_2)$. 
\end{definition}
%
Using the Kantorovich-Rubenstein duality~\cite{villani2008optimal}, WD can be written in the dual form:
\begin{align} 
    \mathcal{W}(\pv_1, \pv_2) = \sup_{||g||_L \leq 1} \mathbb{E}_{\xv_1\sim \pv_1}\left[ g(\bm \xv_1) \right] - \mathbb{E}_{\xv_2\sim \pv_2}\left[ g(\bm \xv_2) \right]\,, \nonumber
\end{align}
where $g$ is a function (often parameterized as a neural network) satisfying the $1$-Lipschitz constraint (\emph{i.e.}, $||g||_L \leq 1$).

\section{Method}


We present the proposed Wasserstein learning framework for KD, where ($i$) the dual form is used for global contrastive knowledge transfer (Sec.~\ref{subsec:wasserstein_dual_form}), and ($ii$) the primal form is adopted for local contrastive knowledge transfer (Sec.~\ref{subsec:wasserstein_primal_form}). The full algorithm is summarized in Sec.~\ref{sec:final_algorithm}.

\subsection{Global Contrastive Knowledge Transfer}\label{subsec:wasserstein_dual_form}

For \emph{global} knowledge transfer, we consider maximizing the mutual information (MI) between feature representations $\hv^S, \hv^T$ at the penultimate layer (before logits) from the teacher and student networks. That is, we seek to match the joint distribution $\pv(\hv^T, \hv^S)$ with the product of marginal distributions $\muv(\hv^T)$ and $\nuv(\hv^S)$ via KL divergence:
\begin{equation}
\label{eq:MI}
    I(\hv^T; \hv^S) = \mathrm{KL}(\pv(\hv^T, \hv^S) \| \muv(\hv^T)\nuv(\hv^S))\,.
\end{equation}
%
Since both the joint and marginal distributions are implicit, (\ref{eq:MI}) cannot be computed directly. To approximate the MI, Noise Contrastive Estimation (NCE)~\cite{gutmann2010noise} is used. Specifically,
we denote a \textit{congruent} pair as one drawn from the joint distribution, and an \textit{incongruent} pair as one drawn independently from the product of marginal distributions.
In other words, a congruent pair is one where the same data input is fed to both the teacher and student networks, while an incongruent pair consists of different data inputs. 
We then define a distribution $\qv$ conditioned on $\eta$ that captures whether the pair is congruent ($\qv(\eta=1)$) or incongruent ($\qv(\eta=0)$), with 
\begin{align}
    \qv(\hv^T, \hv^S|\eta=1) &= \pv(\hv^T, \hv^S)\,, \\ \qv(\hv^T, \hv^S|\eta=0) &=\muv(\hv^T)\nuv(\hv^S)\,.
\end{align}
With one congruent and one incongruent pair,
the prior on $\eta$ is 
\begin{align}
    \qv(\eta=1) = \qv(\eta=0) =1/(1+1)=1/2 \,. 
\end{align}
By Bayes' rule, we can obtain the posterior for $\eta=1$:
%
\begin{align}
\label{eq:bayes}
\qv(\eta=1 | \hv^T, \hv^S) =
\frac{\pv(\hv^T, \hv^S)}{\pv(\hv^T, \hv^S) + \muv(\hv^T)\nuv(\hv^S)}\,,
\end{align}
which can be connected with MI via the following:
\begin{equation}
    \log \qv(\eta=1|\hv^T, \hv^S) \leq  \log \frac{\pv(\hv^T, \hv^S)}{\muv(\hv^T)\nuv(\hv^S)}\,.
\end{equation}
By taking the expectation of both sides w.r.t. the joint distribution $\pv(\hv^T,\hv^S)$, we have:
%
\begin{align}
\label{eq:MI_lowerbound}
    I(\hv^T, \hv^S) \geq \Ep_{\qv(\hv^T, \hv^S|\eta=1)} [\log \qv(\eta=1 | \hv^T, \hv^S)]\,.
\end{align}

We can then maximize the right hand side of (\ref{eq:MI_lowerbound}) to increase the lower bound of the MI.
Though there is no closed form for $\qv(\eta=1 | \hv^T, \hv^S)$, a neural network $g$ (called a \emph{critic} with parameters $\phiv$) can be used to estimate whether a pair comes from the joint distribution or the marginals. 
This shares a similar role as the discriminator of a Generative Adversarial Network (GAN)~\cite{goodfellow2014generative}.
The critic $g$ can be learned via the following NCE loss:
\begin{align}
    \label{eq:nce}
    \Lcal_{\mathrm{NCE}} =&  \Ep_{\qv(\hv^T, \hv^S|\eta=1)}[\log g(\hv^T, \hv^S)] \nonumber\\
    &+ \Ep_{\qv(\hv^T, \hv^S|\eta=0)}[\log(1 - g(\hv^T, \hv^S))] \,.
\end{align}
The parameters $\thetav_S$ and $\phiv$ can be optimized jointly by maximizing (\ref{eq:nce}).

\begin{figure}[!t]
    \centering
    \includegraphics[width=0.5\textwidth]{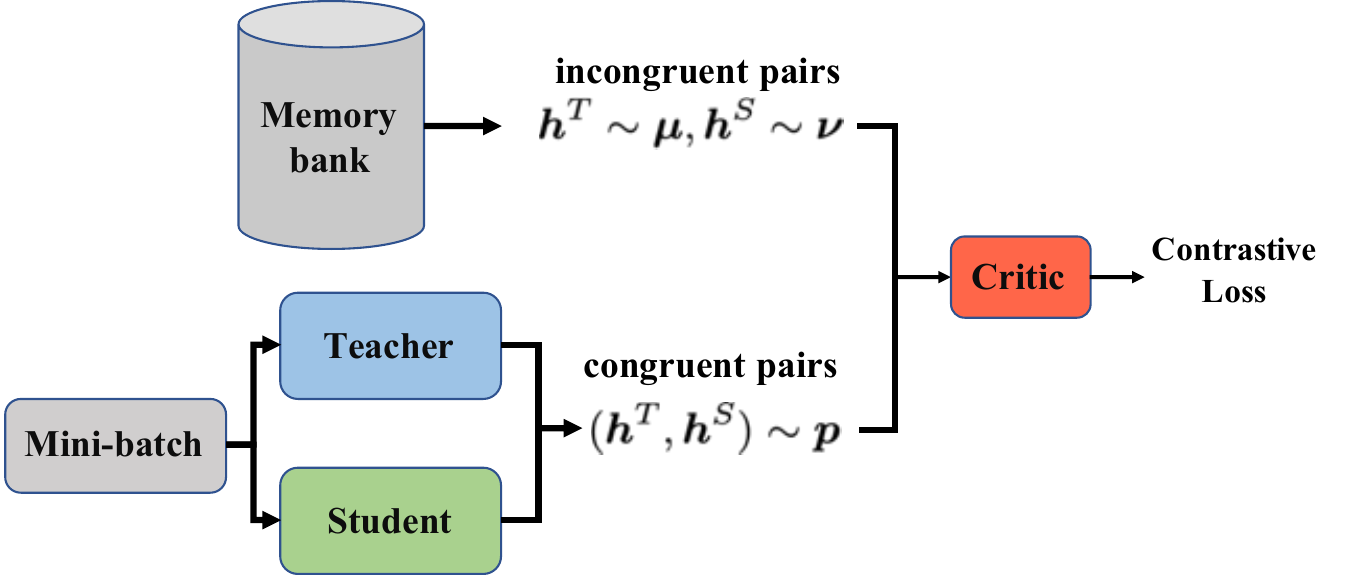}
    \vspace{-5mm}
    \caption{\small\label{fig:wnce_framework} Illustration of Global Contrastive Knowledge Transfer (GCKT) via the use of the dual form for Wasserstein distance.}
    \vspace{-3mm}
\end{figure}

In previous work~\cite{tian2019contrastive}, the critic $g$ is a neural network that maps $(\hv^T, \hv^S)$ to $[0,1]$ without other constraints. This can suffer from several drawbacks: ($i$) $g$ could be sensitive to small numerical changes in the input~\cite{tian2019contrastive, ozair2019wasserstein}, yielding poor performance, especially when the network architectures or training datasets for the student and teacher networks are different. 
($ii$) It can suffer from mode collapse, as the support for $\pv(\hv^T,\hv^S)$ and $\muv(\hv^T) \nuv(\hv^S)$ may not overlap~\cite{wgan}. To alleviate these issues,
we propose using the dual form of Wasserstein distance, by reformulating
(\ref{eq:nce}) as: 
\begin{align}
\label{eq:wd_nce}
    \Lcal_{\mathrm{GCKT}}(\thetav_S,\phiv) =& \, \Ep_{\qv(\hv^T, \hv^S|\eta=1)}[ \hat{g}(\hv^T, \hv^S)] \nonumber \\
    &- \Ep_{\qv(\hv^T, \hv^S|\eta=0)}[ \hat{g}(\hv^T, \hv^S)]\,,
\end{align}
where the new critic function $\hat{g}$ has to satisfy the 1-Lipschitz constraint.
Equation (\ref{eq:wd_nce}) is otherwise similar to (\ref{eq:nce}), 
which not only serves as a lower bound for the mutual information between the student and teacher representations, but also provides a robust critic to better match $\pv(\hv^T,\hv^S)$ with $\muv(\hv^T)\nuv(\hv^S)$. 

Instead of enforcing 1-Lipschitz with the gradient penalty as in \cite{gulrajani2017improved}, we apply spectral normalization \cite{miyato2018spectral} to the critic $\hat{g}$. Specifically,
spectral normalization on an arbitrary matrix $\Amat$ is defined as $\sigma(\Amat)=\max_{\|\betav\|_2\leq 1} \|\Amat\betav\|_2$, which is equivalent to the largest singular value of $\Amat$. 
By applying this regularizer to the weights of each layer in $\hat{g}$, the 1-Lipschitz constraint can be enforced.

Note that when multiple incongurent pairs are chosen, the prior distribution on $\eta$ will also change, and (\ref{eq:wd_nce}) will be updated accordingly to:
\begin{align}
\footnotesize
\label{eq:wnce}
    \Lcal_{\mathrm{GCKT}} (\thetav_S, \phiv) =& \, \Ep_{\qv(\hv^T, \hv^S|\eta=1)}[ \hat{g}(\hv^T, \hv^S)] \\
    &- M  \Ep_{\qv(\hv^T, \hv^S|\eta=0)}[ \hat{g}(\hv^T, \hv^S)] \,, \nonumber
\end{align}
where $M>1$, and the lower bound for the mutual information will be tightened with large $M$~\cite{gutmann2010noise, tian2019contrastive}.
In practice, the incongruent samples are drawn from a memory buffer, that stores pre-computed features of every data sample from previous mini-batches. In this way, we are able to efficiently retrieve a large number of negative samples without recalculating the features. 
Due to the use of data samples across multiple mini-batches for Wasserstein constrastive learning, we denote this method as \emph{global} contrastive knowledge transfer (GCKT), as illustrated in Figure \ref{fig:wnce_framework}.



\subsection{Local Contrastive Knowledge Transfer}\label{subsec:wasserstein_primal_form}
\begin{figure}[!t]
    \centering
    \includegraphics[width=0.45\textwidth]{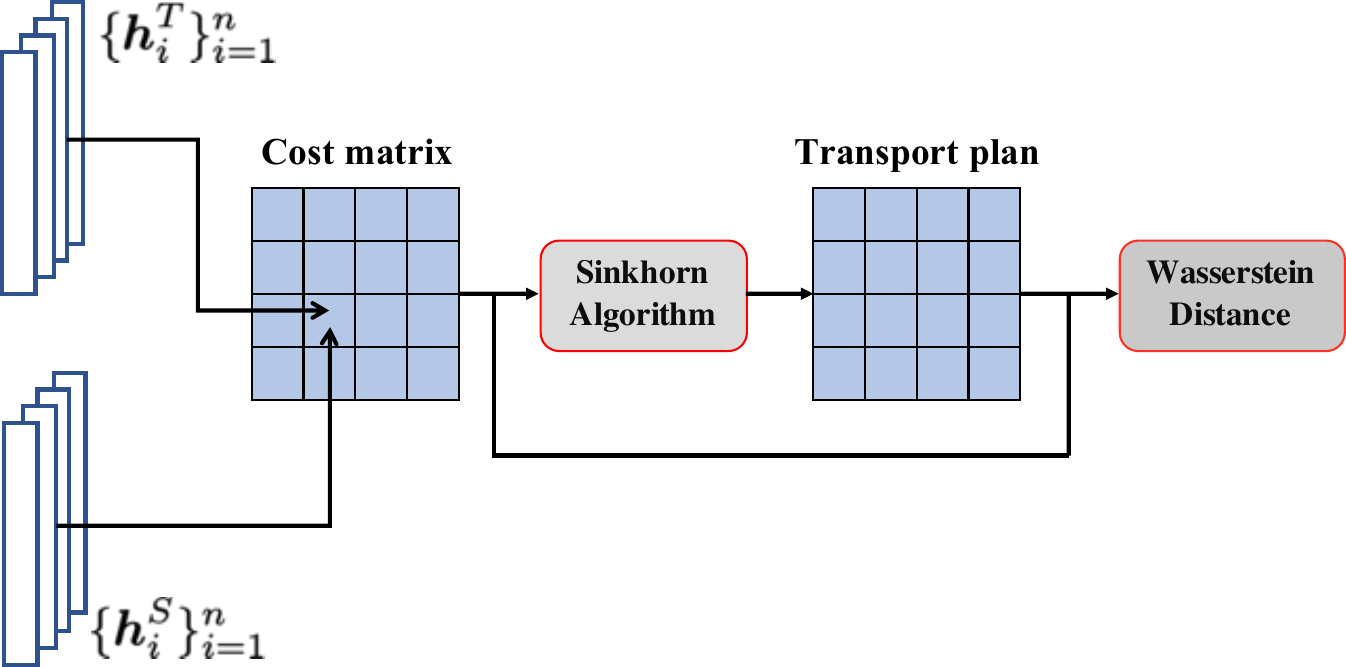}
    \vspace{-2mm}
    \caption{\small\label{fig:pwd_framework} Illustration of Local Contrastive Knowledge Transfer (LCKT) via the use of the primal form for Wasserstein distance.}
    \vspace{-4mm}
\end{figure}
Contrastive learning can also be applied within a mini-batch to further enhance performance. Specifically, in a mini-batch, the features $\{\hv^T_i\}_{i=1}^n$ extracted from the teacher network can be viewed as a fixed set when training the student network. Ideally, categorical information is encapsulated in the feature space, so each element $\{\hv^S_j\}_{j=1}^n$ from the student network should be able to find close neighbors in this set. For instance, nearby samples may share the same class. Therefore, we encourage the model to push $\hv^S_j$ to several neighbors $\{\hv^T_i\}_{i=1}^n$  instead of just one from the teacher network for better generalization. As the distribution matching happens in a mini-batch, we denote this as \emph{local} contrastive knowledge transfer (LCKT).

This can be implemented efficiently with the primal form of Wasserstein distance. 
Specifically, the primal form can be interpreted as a less expensive way to transfer probability mass from $\muv(\hv^T)$ to $\nuv(\hv^S)$, when only finite training samples are used. That is, we have $\muv(\hv^T) = \sum_{i=1}^n \ervu_i\delta_{\hv^T_i}$, $\nuv(\hv^S) = \sum_{j=1}^n \ervv_j\delta_{\hv^S_j}$, where $\delta_{\xv}$ is the Dirac function centered on $\xv$. Since $\muv(\hv^T), \nuv(\hv^S)$ are valid probability distributions, $\uv=\{\ervu_i\}_{i=1}^n, \vv=\{\ervv_j\}_{j=1}^n$ both lie on a simplex, \emph{i.e.}, $\sum_{i=1}^n \ervu_i = 1$, and $\sum_{j=1}^n \ervv_j = 1$. 
Under this setting, 
the primal form can be reformulated into:
\begin{align}\label{eq:wasserstein_primal_finite}
    & \mathcal{W}(\muv, \nuv) = \min_{\piv} \sum_{i=1}^n \sum_{j=1}^n \piv_{ij} c(\hv^T_i, \hv^S_j) = \min_{\piv} \langle\piv, \Cmat \rangle \nonumber\\
    & \mathrm{s.t. \quad} \sum_{j=1}^n \piv_{ij} = \ervu_i, \quad \sum_{i=1}^n \piv_{ij} = \ervv_j\,,
\end{align}
where $\piv$ is the discrete joint probability in $\hv^T$ and $\hv^S$ (\emph{i.e.}, the transport plan), $\Cmat$ is the cost matrix given by $\Cmat_{ij}=c(\hv_i^T,\hv_j^S)$, and $\langle \piv, \Cmat \rangle = \mathrm{Tr}(\piv^\top \Cmat)$ represents the Frobenius dot-product. Expression $c(\cdot)$ is a cost function measuring the dissimilarity between the two feature vectors, where cosine distance $c(\xv,\yv)=1-\frac{\xv^\top\yv}{||\xv||_2||\yv||_2}$ is a popular choice.
Ideally, the global optimum for (\ref{eq:wasserstein_primal_finite}) may be obtained using linear programming~\cite{villani2008optimal, peyre2017computational}. However, this method is not differentiable, making it incompatible with existing deep learning frameworks. 
As an alternative, the Sinkhorn algorithm~\cite{cuturi2013sinkhorn} is applied to solve (\ref{eq:wasserstein_primal_finite}) by adding a convex regularization term, \emph{i.e.,}
\begin{align}\label{eq:sinkhorn_finite}
    & \mathcal{L}_{\mathrm{LCKT}}(\thetav_S) = \min_{\piv} \sum_{i,j} \piv_{ij} c(\hv^T_i, \hv^S_j) + \epsilon \mathcal{H}(\piv)\,,
\end{align}
where $\mathcal{H}(\bm \pi) = \sum_{i,j} \piv_{ij} \log \piv_{ij}$, and $\epsilon$ is the hyper-parameter controlling the importance of the entropy loss on $\piv$. 
Detailed procedures for solving this is summarized in Algorithm~\ref{alg:sink}.
\begin{algorithm}[!t]
\caption{Sinkhorn Algorithm.}
\label{alg:sink}
\begin{algorithmic}[1]
\STATE {\bfseries Input:} \footnotesize{ \{$\hv^T_i\}_{i=1}^n$,$\{\hv^S_j\}_{j=1}^n$, $\epsilon$, probability vectors $\muv$, $\nuv$}
\STATE $\boldsymbol{\sigma}=\frac{1}{n}\mathbf{1_n}$, $\boldsymbol{\pi}^{(1)} = \mathbf{1} \mathbf{1}^\top$
\STATE $\Cmat_{ij} = c(\hv^T_i, \hv^S_j)$, $\Amat_{ij} = {\rm e}^{-\frac{\Cmat_{ij}}{\epsilon}}$
\FOR{$t=1,2,3\ldots$}
    \STATE $\Qmat = \Amat \odot \boldsymbol{\pi}^{(t)}$ \footnotesize{// $\odot$ is Hadamard product}
    \FOR{$k=1,2,3,\ldots K$}
        \STATE $\boldsymbol{\delta} = \frac{\muv}{n\Qmat{\boldsymbol{\sigma}}}$, $\boldsymbol{\sigma} = \frac{\nuv}{n\Qmat^\top\boldsymbol{\delta}}$
    \ENDFOR
    \STATE $\boldsymbol{\pi}^{(t+1)} = \text{diag}(\boldsymbol{\delta})\Qmat\text{diag}(\boldsymbol{\sigma})$
\ENDFOR
\STATE $\Wcal=\langle \boldsymbol{\pi}, \Cmat \rangle$  // \footnotesize{$\langle \cdot, \cdot \rangle$ is the Frobenius dot-product} 
\STATE Return $\Wcal$

\end{algorithmic}
\end{algorithm}
Although Lines $4$-$10$ in Algorithm~\ref{alg:sink} constitute an iterative algorithm, its time complexity is small compared to the other feed-forward modules. Also, thanks to the Envelop Theorem~\cite{carter2001foundations}, we can ignore the gradient flow through $\piv$, meaning that there is no need to back-propagate gradients for Lines $4$-$10$. In practice, we can simply detach/stop-gradient the for-loop module in Pytorch or Tensorflow, while the loss can still help refine the feature representations. 
Figure \ref{fig:pwd_framework} illustrates the procedure for calculating the Wasserstein distance.

\subsection{Unifying Global and Local Knowledge Transfer}

\label{sec:final_algorithm}
Global knowledge transfer is designed for matching the joint distribution $\pv(\hv_T, \hv_S)$ with the product of the marginal distributions $\muv(\hv_T)\nuv(\hv_S)$ via contrastive learning under a Wasserstein metric, achieving better distillation.
At the same time, local knowledge transfer incentivizes matching the marginal distribution $\muv(\hv_T)$ with $\nuv(\hv_S)$ via optimal transport, aiming for better generalization. 
Section \ref{subsec:wasserstein_dual_form} optimizes the MI by maximizing the lower bound, while Section \ref{subsec:wasserstein_primal_form} minimizes (\ref{eq:sinkhorn_finite}) to match the feature distributions.

Although GCKT and LCKT are designed for different objectives, they are complementary to each other. By optimizing LCKT, we aim to minimize the discrepancy between the marginal distributions, which is equivalent to reducing the difference between the two feature spaces, so that LCKT can provide a more constrained feature space for GCKT. On the other hand,
by optimizing GCKT, the learned representation can also form a better feature space, which in turn helps LCKT match the marginal distributions.

In summary, the training objective for our method is written as follows:
\begin{align}
\label{eq:wcord}
    \Lcal_{\mathrm{WCoRD}} (\thetav_S,\phiv) =& \, \Lcal_{\mathrm{CE}}(\thetav_S)-\lambda_1\Lcal_{\mathrm{GCKT}}(\thetav_S,\phiv) \nonumber\\
    &+ \lambda_2 \Lcal_{\mathrm{LCKT}}(\thetav_S) \,,
\end{align}
where besides the parameters $\thetav_S$ of the student network, an additional set of parameters $\phiv$ for the critic is also learned.

\section{Related Work}

\noindent\textbf{Knowledge Distillation}\,
Recent interest in knowledge distillation can be traced back to \cite{hinton2015distilling}, though similar ideas have been proposed before~\cite{zeng2000using, bucilua2006model}. 
These methods use the probability distribution of the output over the prediction classes of a large teacher network as additional supervision signals to train a smaller student network. 
Recent studies have suggested alternative distillation objectives.
Later works such as FitNet~\cite{romero2014fitnets} extend the idea by using the intermediate layers instead of only $\zv^T$. \cite{zagoruyko2016paying} proposed to use an attention map transfer in KD. SPKD~\cite{tung2019similarity} also utilizes intermediate features, but tries to mimic the representation space of the teacher features, rather than preserving pairwise similarities like FitNet.
More recently, Contrastive Representation Distillation (CRD)~\cite{tian2019contrastive} proposed applying NCE~\cite{gutmann2010noise} to an intermediate layer.
Another line of KD research explores alternatives to the teacher-student training paradigm. For example,
\cite{zhu2018knowledge} proposed an on-the-fly ensemble teacher network, in which the teacher is jointly trained with multiple students under a multi-branch network architecture, and the teacher's prediction is a weighted average of predictions from all the branches. Most recenlty, \cite{yuan2020revisiting} shows that KD can be understood as label smoothing regularization. 

\begin{table*}[t!]
\small
\begin{center}
\begin{tabular}{lccccccc}
\toprule
\thead{Teacher \\ Student} & \thead{WRN-40-2 \\ WRN-16-2} & \thead{WRN-40-2 \\ WRN-40-1} & \thead{resnet56\\resnet20} & \thead{resnet110\\resnet20} & \thead{resnet110\\resnet32} & \thead{resnet32x4\\resnet8x4} & \thead{vgg13\\vgg8}\\
\midrule
Teacher & 75.61 & 75.61 & 72.34 & 74.31 & 74.31 & 79.42 & 74.64 \\
Student & 73.26 & 71.98 & 69.06 & 69.06 & 71.14 & 72.50 & 70.36 \\
\midrule

KD  & 74.92 & 73.54 & 70.66 & 70.67 & 73.08 & 73.33 & 72.98 \\
FitNet  & 73.58 ($\mathcolor{red}{\downarrow}$)& 72.24 ($\mathcolor{red}{\downarrow}$)& 69.21 ($\mathcolor{red}{\downarrow}$)& 68.99 ($\mathcolor{red}{\downarrow}$)& 71.06 ($\mathcolor{red}{\downarrow}$)& 73.50 ($\mathcolor{green}{\uparrow}$)& 71.02 ($\mathcolor{red}{\downarrow}$)\\
AT & 74.08 ($\mathcolor{red}{\downarrow}$)& 72.77 ($\mathcolor{red}{\downarrow}$)& 70.55 ($\mathcolor{red}{\downarrow}$)& 70.22 ($\mathcolor{red}{\downarrow}$)& 72.31 ($\mathcolor{red}{\downarrow}$)& 73.44 ($\mathcolor{green}{\uparrow}$)& 71.43 ($\mathcolor{red}{\downarrow}$)\\
SP & 73.83 ($\mathcolor{red}{\downarrow}$)& 72.43 ($\mathcolor{red}{\downarrow}$)& 69.67 ($\mathcolor{red}{\downarrow}$)& 70.04 ($\mathcolor{red}{\downarrow}$)& 72.69 ($\mathcolor{red}{\downarrow}$)& 72.94 ($\mathcolor{red}{\downarrow}$)& 72.68 ($\mathcolor{red}{\downarrow}$)\\
CC & 73.56 ($\mathcolor{red}{\downarrow}$)& 72.21 ($\mathcolor{red}{\downarrow}$)& 69.63 ($\mathcolor{red}{\downarrow}$)& 69.48 ($\mathcolor{red}{\downarrow}$)& 71.48 ($\mathcolor{red}{\downarrow}$)& 72.97 ($\mathcolor{red}{\downarrow}$)& 70.71 ($\mathcolor{red}{\downarrow}$)\\
VID & 74.11 ($\mathcolor{red}{\downarrow}$)& 73.30 ($\mathcolor{red}{\downarrow}$)& 70.38 ($\mathcolor{red}{\downarrow}$)& 70.16 ($\mathcolor{red}{\downarrow}$)& 72.61 ($\mathcolor{red}{\downarrow}$)& 73.09 ($\mathcolor{red}{\downarrow}$)& 71.23 ($\mathcolor{red}{\downarrow}$)\\
RKD & 73.35 ($\mathcolor{red}{\downarrow}$)& 72.22 ($\mathcolor{red}{\downarrow}$)& 69.61 ($\mathcolor{red}{\downarrow}$)& 69.25 ($\mathcolor{red}{\downarrow}$)& 71.82 ($\mathcolor{red}{\downarrow}$)& 71.90 ($\mathcolor{red}{\downarrow}$)& 71.48 ($\mathcolor{red}{\downarrow}$)\\
PKT & 74.54 ($\mathcolor{red}{\downarrow}$)& 73.45 ($\mathcolor{red}{\downarrow}$)& 70.34 ($\mathcolor{red}{\downarrow}$)& 70.25 ($\mathcolor{red}{\downarrow}$)& 72.61 ($\mathcolor{red}{\downarrow}$)& 73.64 ($\mathcolor{green}{\uparrow}$)& 72.88 ($\mathcolor{red}{\downarrow}$)\\
AB & 72.50 ($\mathcolor{red}{\downarrow}$)& 72.38 ($\mathcolor{red}{\downarrow}$)& 69.47 ($\mathcolor{red}{\downarrow}$)& 69.53 ($\mathcolor{red}{\downarrow}$)& 70.98 ($\mathcolor{red}{\downarrow}$)& 73.17 ($\mathcolor{red}{\downarrow}$)& 70.94 ($\mathcolor{red}{\downarrow}$)\\
FT  & 73.25 ($\mathcolor{red}{\downarrow}$)& 71.59 ($\mathcolor{red}{\downarrow}$)& 69.84 ($\mathcolor{red}{\downarrow}$)& 70.22 ($\mathcolor{red}{\downarrow}$)& 72.37 ($\mathcolor{red}{\downarrow}$)& 72.86 ($\mathcolor{red}{\downarrow}$)& 70.58 ($\mathcolor{red}{\downarrow}$)\\
FSP & 72.91 ($\mathcolor{red}{\downarrow}$)& 
 n/a & 69.95 ($\mathcolor{red}{\downarrow}$)& 70.11 ($\mathcolor{red}{\downarrow}$)& 71.89 ($\mathcolor{red}{\downarrow}$)& 72.62 ($\mathcolor{red}{\downarrow}$)& 70.23 ($\mathcolor{red}{\downarrow}$)\\
NST  & 73.68 ($\mathcolor{red}{\downarrow}$)& 72.24 ($\mathcolor{red}{\downarrow}$)& 69.60 ($\mathcolor{red}{\downarrow}$)& 69.53 ($\mathcolor{red}{\downarrow}$)& 71.96 ($\mathcolor{red}{\downarrow}$)& 73.30 ($\mathcolor{red}{\downarrow}$)& 71.53 ($\mathcolor{red}{\downarrow}$)\\
CRD & {75.48} ($\mathcolor{green}{\uparrow}$)& {74.14} ($\mathcolor{green}{\uparrow}$)& {71.16} ($\mathcolor{green}{\uparrow}$)& {71.46} ($\mathcolor{green}{\uparrow}$)& {73.48} ($\mathcolor{green}{\uparrow}$)& {75.51} ($\mathcolor{green}{\uparrow}$)& {73.94} ($\mathcolor{green}{\uparrow}$)\\
CRD+KD & 75.64 ($\mathcolor{green}{\uparrow}$)& 74.38 ($\mathcolor{green}{\uparrow}$)& 71.63 ($\mathcolor{green}{\uparrow}$)& 71.56 ($\mathcolor{green}{\uparrow}$)& 73.75 ($\mathcolor{green}{\uparrow}$)& 75.46 ($\mathcolor{green}{\uparrow}$)& 74.29 ($\mathcolor{green}{\uparrow}$)\\
\midrule
LCKT & 75.22 ($\mathcolor{green}{\uparrow}$)& 74.11 ($\mathcolor{green}{\uparrow}$)& 71.14 ($\mathcolor{green}{\uparrow}$)& 71.23 ($\mathcolor{green}{\uparrow}$)& 72.32 ($\mathcolor{green}{\uparrow}$)& 74.65 ($\mathcolor{green}{\uparrow}$)& 73.50 ($\mathcolor{green}{\uparrow}$)\\

GCKT & 75.47 ($\mathcolor{green}{\uparrow}$)& 74.23 ($\mathcolor{green}{\uparrow}$)& 71.21 ($\mathcolor{green}{\uparrow}$)& 71.43 ($\mathcolor{green}{\uparrow}$)& 73.41 ($\mathcolor{green}{\uparrow}$)& 75.45 ($\mathcolor{green}{\uparrow}$)& 74.10 ($\mathcolor{green}{\uparrow}$)\\

WCoRD & 75.88 ($\mathcolor{green}{\uparrow}$)& \textbf{74.73} ($\mathcolor{green}{\uparrow}$)& 71.56 ($\mathcolor{green}{\uparrow}$)& 71.57 ($\mathcolor{green}{\uparrow}$)& 73.81 ($\mathcolor{green}{\uparrow}$)& 75.95 ($\mathcolor{green}{\uparrow}$)& 74.55 ($\mathcolor{green}{\uparrow}$)\\
                                                      
WCoRD+KD & \textbf{76.11} ($\mathcolor{green}{\uparrow}$)& {74.72} ($\mathcolor{green}{\uparrow}$)& \textbf{71.92} ($\mathcolor{green}{\uparrow}$)& \textbf{71.88} ($\mathcolor{green}{\uparrow}$)& \textbf{74.20} ($\mathcolor{green}{\uparrow}$)& \textbf{76.15} ($\mathcolor{green}{\uparrow}$)& \textbf{74.72} ($\mathcolor{green}{\uparrow}$)\\
\bottomrule
\end{tabular}
\caption{
\small{
CIFAR-100 test \emph{accuracy} (\%) of student networks trained with a number of distillation methods, when sharing the same architecture type as the teacher. See Appendix for citations of the compared methods. $\mathcolor{green}{\uparrow}$ denotes outperformance over KD, and $\mathcolor{red}{\downarrow}$ denotes underperformance. For all other methods, we used author-provided or author-verified code from the CRD repository. Our reported results are averaged over 5 runs. Note that $\lambda_1=0.8$ is the same as the weight on CRD, and $\lambda_2=0.05$.
}}
\label{tb:cifar100_same}
\end{center}
\vspace{-2mm}
\end{table*}

\begin{table*}[t]
\small

\begin{center}
\resizebox{\linewidth}{!}{
\begin{tabular}{l|cc|cccccc|cccc}
\toprule
 & Teacher & Student & AT & KD & SP & CC & CRD & CRD+KD & LCKT & GCKT & WCoRD & WCoRD+KD \\
\midrule
Top-1 & 26.69 & 30.25 & 29.30 & 29.34 & 29.38 & 30.04 & 28.83 & 28.62 & 29.10 & 28.78 & 28.51 & \textbf{28.44}\\
Top-5 & 8.58 & 10.93 & 10.00 & 10.12 & 10.20 & 10.83 & 9.87 & 9.51 & 10.05 & 9.92 & 9.84 & \textbf{9.45}\\
\bottomrule
\end{tabular}
}
\caption{
\small{Top-1 and Top-5 error rates (\%) of student network ResNet-18 on ImageNet validation set.}
}
\label{tb:imagenet}
\end{center}
\vspace{-3mm}
\end{table*}

\vspace{5pt}
\noindent\textbf{Optimal Transport}\,
Optimal transport distance, \emph{a.k.a.} Wasserstein distance, has a long history in mathematics, with applications ranging from resource allocation to computer vision~\cite{rubner1998metric}. Traditionally, optimal transport problems are solved by linear/quadratic programming.
Within deep learning, the \emph{dual} form of the Wasserstein distance is used by \cite{wgan, gulrajani2017improved} as an alternative metric for distribution matching in Generative Adversarial Network (GAN) \cite{goodfellow2014generative}, where the dual form is approximated by imposing a 1-Lipschitz constraint on the critic. 
The \emph{primal} form of Wasserstein distance can be solved by the Sinkhorn algorithm~\cite{cuturi2013sinkhorn}, which has been applied to a wide range of deep learning tasks, including document retrieval and classification~\cite{kusner2015word}, sequence-to-sequence learning~\cite{chen2019improving}, adversarial attacks~\cite{wong2019wasserstein}, graph matching~\cite{xu2019gromov}, and cross-domain alignment~\cite{chen2020graph}.
To the authors' knowledge, this work is the first to apply optimal transport to KD, and to utilize both its primal and dual forms to construct a general Wasserstein learning framework. 


\vspace{5pt}
\noindent\textbf{Contrastive Learning} \,
Contrastive learning~\cite{gutmann2010noise,arora2019theoretical} is a popular research area that has been successfully applied to density estimation and representation learning, especially in self-supervised setting~\cite{he2019momentum,chen2020simple}. It has been shown that the contrastive objective can be interpreted as maximizing the lower bound of mutual information between different views of the data~\cite{hjelm2018learning,oord2018representation,bachman2019learning,henaff2019data}, though it remains unclear whether the success is determined by mutual information or by the specific form of the contrastive loss~\cite{tschannen2019mutual}.  Recently, contrastive learning has been extended to Wasserstein dependency measure~\cite{ozair2019wasserstein}. Our global  contrastive transfer shares similar ideas with it. However, its application to KD has not been studied before. Further, both the primal and dual forms are used to form an integral framework.

\section{Experiments}
We evaluate the proposed WCoRD framework on three knowledge distillation tasks: ($i$) model compression of a large network, ($ii$) cross-modal transfer, and ($iii$) privileged information distillation. 

\begin{table*}[t!]
\small
\begin{center}
\begin{tabular}{lcccccc}
\toprule
\thead{Teacher \\ Student} & \thead{vgg13 \\ MobileNetV2} & \thead{ResNet50 \\ MobileNetV2} & \thead{ResNet50 \\ vgg8} & \thead{resnet32x4\\ShuffleNetV1} & \thead{resnet32x4\\ShuffleNetV2} & \thead{WRN-40-2\\ShuffleNetV1}\\
\midrule
Teacher & 74.64	& 79.34 & 79.34	& 79.42	& 79.42	& 75.61 \\
Student & 64.6 & 64.6 & 70.36 & 70.5 & 71.82 & 70.5 \\
\midrule

KD  & 67.37 & 67.35 & 73.81 & 74.07 & 74.45 & 74.83 \\
FitNet  & 64.14 ($\mathcolor{red}{\downarrow}$)& 63.16 ($\mathcolor{red}{\downarrow}$)& 70.69 ($\mathcolor{red}{\downarrow}$)& 73.59 ($\mathcolor{red}{\downarrow}$)& 73.54 ($\mathcolor{red}{\downarrow}$)& 73.73 ($\mathcolor{red}{\downarrow}$)\\
AT & 59.40 ($\mathcolor{red}{\downarrow}$)& 58.58 ($\mathcolor{red}{\downarrow}$)& 71.84 ($\mathcolor{red}{\downarrow}$)& 71.73 ($\mathcolor{red}{\downarrow}$)& 72.73 ($\mathcolor{red}{\downarrow}$)& 73.32 ($\mathcolor{red}{\downarrow}$)\\
SP & 66.30 ($\mathcolor{red}{\downarrow}$)& 68.08 ($\mathcolor{green}{\uparrow}$)& 73.34 ($\mathcolor{red}{\downarrow}$)& 73.48 ($\mathcolor{red}{\downarrow}$)& 74.56 ($\mathcolor{green}{\uparrow}$)& 74.52 ($\mathcolor{red}{\downarrow}$)\\
CC & 64.86 ($\mathcolor{red}{\downarrow}$)& 65.43 ($\mathcolor{red}{\downarrow}$)& 70.25 ($\mathcolor{red}{\downarrow}$)& 71.14 ($\mathcolor{red}{\downarrow}$)& 71.29 ($\mathcolor{red}{\downarrow}$)& 71.38 ($\mathcolor{red}{\downarrow}$)\\
VID & 65.56 ($\mathcolor{red}{\downarrow}$)& 67.57 ($\mathcolor{green}{\uparrow}$)& 70.30 ($\mathcolor{red}{\downarrow}$)& 73.38 ($\mathcolor{red}{\downarrow}$)& 73.40 ($\mathcolor{red}{\downarrow}$)& 73.61 ($\mathcolor{red}{\downarrow}$)\\
RKD & 64.52 ($\mathcolor{red}{\downarrow}$)& 64.43 ($\mathcolor{red}{\downarrow}$)& 71.50 ($\mathcolor{red}{\downarrow}$)& 72.28 ($\mathcolor{red}{\downarrow}$)& 73.21 ($\mathcolor{red}{\downarrow}$)& 72.21 ($\mathcolor{red}{\downarrow}$)\\
PKT & 67.13 ($\mathcolor{red}{\downarrow}$)& 66.52 ($\mathcolor{red}{\downarrow}$)& 73.01 ($\mathcolor{red}{\downarrow}$)& 74.10 ($\mathcolor{green}{\uparrow}$)& 74.69 ($\mathcolor{green}{\uparrow}$)& 73.89 ($\mathcolor{red}{\downarrow}$)\\
AB & 66.06 ($\mathcolor{red}{\downarrow}$)& 67.20 ($\mathcolor{red}{\downarrow}$)& 70.65 ($\mathcolor{red}{\downarrow}$)& 73.55 ($\mathcolor{red}{\downarrow}$)& 74.31 ($\mathcolor{red}{\downarrow}$)& 73.34 ($\mathcolor{red}{\downarrow}$)\\
FT  & 61.78 ($\mathcolor{red}{\downarrow}$)& 60.99 ($\mathcolor{red}{\downarrow}$)& 70.29 ($\mathcolor{red}{\downarrow}$)& 71.75 ($\mathcolor{red}{\downarrow}$)& 72.50 ($\mathcolor{red}{\downarrow}$)& 72.03 ($\mathcolor{red}{\downarrow}$)\\
NST  & 58.16 ($\mathcolor{red}{\downarrow}$)& 64.96 ($\mathcolor{red}{\downarrow}$)& 71.28 ($\mathcolor{red}{\downarrow}$)& 74.12 ($\mathcolor{green}{\uparrow}$)& 74.68 ($\mathcolor{green}{\uparrow}$)& 74.89 ($\mathcolor{green}{\uparrow}$)\\
CRD & {69.73} ($\mathcolor{green}{\uparrow}$)& {69.11} ($\mathcolor{green}{\uparrow}$)& {74.30} ($\mathcolor{green}{\uparrow}$)& {75.11} ($\mathcolor{green}{\uparrow}$)& {75.65} ($\mathcolor{green}{\uparrow}$)& {76.05} ($\mathcolor{green}{\uparrow}$)\\
CRD+KD & 69.94 ($\mathcolor{green}{\uparrow}$)& 69.54 ($\mathcolor{green}{\uparrow}$)& 74.58 ($\mathcolor{green}{\uparrow}$)& 75.12 ($\mathcolor{green}{\uparrow}$)& 76.05 ($\mathcolor{green}{\uparrow}$)& 76.27 ($\mathcolor{green}{\uparrow}$)\\
\midrule

LCKT & 68.21 ($\mathcolor{green}{\uparrow}$)& 68.81 ($\mathcolor{green}{\uparrow}$)& 73.21 ($\mathcolor{green}{\uparrow}$)& 74.62 ($\mathcolor{green}{\uparrow}$)& 74.70 ($\mathcolor{green}{\uparrow}$)& 75.08  ($\mathcolor{green}{\uparrow}$)\\

GCKT & 68.78 ($\mathcolor{green}{\uparrow}$)& 69.20 ($\mathcolor{green}{\uparrow}$)& 74.29 ($\mathcolor{green}{\uparrow}$)& 75.18 ($\mathcolor{green}{\uparrow}$)& 75.78 ($\mathcolor{green}{\uparrow}$)& 76.13  ($\mathcolor{green}{\uparrow}$)\\

WCoRD & 69.47 ($\mathcolor{green}{\uparrow}$)& \textbf{70.45} ($\mathcolor{green}{\uparrow}$)& \textbf{74.86} ($\mathcolor{green}{\uparrow}$)& 75.40 ($\mathcolor{green}{\uparrow}$)& 75.96 ($\mathcolor{green}{\uparrow}$)& 76.32 ($\mathcolor{green}{\uparrow}$)\\
                                                      
WCoRD+KD & \textbf{70.02} ($\mathcolor{green}{\uparrow}$)& {70.12} ($\mathcolor{green}{\uparrow}$)& {74.68} ($\mathcolor{green}{\uparrow}$)& \textbf{75.77} ($\mathcolor{green}{\uparrow}$)& \textbf{76.48} ($\mathcolor{green}{\uparrow}$)& \textbf{76.68} ($\mathcolor{green}{\uparrow}$)\\
\bottomrule
\end{tabular}
\caption{
\small{
CIFAR-100 test \emph{accuracy} (\%) of a student network trained with a number of distillation methods, when the teacher network architecture is significantly different. We use the same codebase from the CRD repository. Our reported results are averaged over 5 runs. Note that $\lambda_1=0.8$ is the same as the weight on CRD, and $\lambda_2=0.05$.
}}
\label{tb:cifar100_diff}
\end{center}
\vspace{-4mm}
\end{table*}

\subsection{Model Compression}

\noindent\textbf{Experiments on CIFAR-100}\,
CIFAR-100 \cite{krizhevsky2009learning} consists of 50K training images (0.5K images per class) and 10K test images.
For fair comparison, we use the public CRD codebase~\cite{tian2019contrastive} in our experiments. 
Two scenarios are considered: ($i$) the student and the teacher share the same network architecture, and ($ii$) different network architectures are used.

Table~\ref{tb:cifar100_same} and \ref{tb:cifar100_diff} present the top-$1$ accuracy from different distillation methods.
In both tables, models using the original KD is a strong baseline, which only CRD and our WCoRD consistently outperform. 
The strong performance of the original KD method is manifested because distillation is performed between low-dimensional probability distributions from the teacher and student networks,
which makes it relatively easy for the student to learn knowledge from the teacher.
However, if knowledge transfer is applied to features from intermediate layers, the numerical scale of features can be different, even when both teacher and student share the same network architecture. As shown in Table \ref{tb:cifar100_diff}, directly applying similarity matching to align teacher and student features even hurts performance. 

WCoRD is a unified framework bridging GCKT and LCKT, which improves the performance of CRD, a current state-of-the-art model. 
When the same network architecture is used for both the teacher and student networks, an average relative improvement\footnote{The relative improvement is defined as $\frac{\text{WCoRD}-\text{CRD}}{\text{CRD} - \text{KD}}$, where the name of each method represents the corresponding accuracy of the student model.} of $48.63\%$ is achieved (derived from Table \ref{tb:cifar100_same}). This performance lift is $43.27\%$ when different network architectures are used (derived from Table \ref{tb:cifar100_diff}).

We can also add the basic KD loss to WCoRD, obtaining an ensemble distillation loss (denoted as WCoRD+KD), similar to~\cite{tian2019contrastive}. In most cases, this ensemble loss can further improve the performance. However, in ResNet50 $\rightarrow$ MobileNetV2 and ResNet50 $\rightarrow$ VGG8, WCoRD still works better than WCoRD+KD.

\vspace{5pt}
\noindent\textbf{Ablation Study}\,
We report results using only global or local contrastive knowledge transfer in Table \ref{tb:cifar100_same} and \ref{tb:cifar100_diff}. 
LCKT performs better than KD but slightly worse than CRD and GCKT, as both
CRD and GCKT are NCE-based algorithms, where negative samples are used to improve performance.
Additionally, GCKT enforces an 1-Lipschitz constraint on the critic function, which includes an extra hyper-parameter. 
Results show that CRD and GCKT have comparable results, and
in some cases, GCKT performs slightly better (\emph{e.g.}, from VGG13 to VGG8).

\begin{table}[t!]
\small
\begin{center}
\resizebox{\linewidth}{!}{
\begin{tabular}{l|cccccccc}
\toprule
 $\lambda_2$ & 0 & 0.01 & 0.03 & 0.05 & 0.08 & 0.1 & 0.2 \\
\midrule
Mean & 75.45 & 75.66 & 75.75    & \textbf{75.95}    & 75.83    & 75.66 & 75.62\\
Std & 0.31 & 0.46    & 0.44    & 0.40    &  0.34     & 0.29    & 0.47\\
\bottomrule
\end{tabular}
}
    \caption{
    \small{CIFAR-100 test accuracy (\%) of student network ResNet-8x4 with different weights on the local knowledge transfer term. The teacher network is ResNet-32x4.}
    }
\label{tb:cifar100_ablation}
\end{center}
\vspace{-4mm}
\end{table}
\begin{table}[t!]
\small
\begin{center}
\begin{tabular}{l|cccc}
\toprule
 Layer  & 1 & 2 & 3 & 4 \\
\midrule
CRD & \textbf{55.0} & 63.64 & 73.76    & 74.75    \\
WCoRD & 54.6 & \textbf{63.70}   & \textbf{74.23}    & \textbf{75.43} \\

\bottomrule
\end{tabular}
    \caption{
    \small{Top-1 Accuracy (\%) on chrominance view of STL-10 testing set with ResNet-18. The modal is distilled on the network trained with the luminance view of Tiny-ImageNet. }
    }
\label{tb:crossmodal}
\end{center}
\vspace{-7mm}
\end{table}
We perform an additional ablation study on the weight of the $\Lcal_{\text{LCKT}}$ loss term (\emph{i.e.}, $\lambda_2$ in (\ref{eq:wcord})). We adjust $\lambda_2$ from $0$ to $0.2$, and set $\lambda_1=0.8$, which is the same as in CRD for fair comparison.
  Results are summarized in Table \ref{tb:cifar100_ablation}. The standard deviation (Std) is reported based on 5 runs.
We observe that:
($i$) when $\lambda_2=0.05$, ResNet-8x4 performs the best; and ($ii$) WCoRD can consistently outperform GCKT and CRD methods, when $\lambda_2\in(0,0.2]$. 

\vspace{5pt}
\noindent\textbf{Experiments on ImageNet}\,
We also evaluate the proposed method on a larger dataset,
ImageNet~\cite{deng2009imagenet}, which contains $1.2$M images for training and $50$K for validation.
In this experiment, we use ResNet-34~\cite{he2016deep} as the teacher and ResNet-18 as the student, and use the same training setup as in CRD~\cite{tian2019contrastive} for fair comparison.
Top-$1$ and Top-$5$ error rates (lower is better) are reported in Table~\ref{tb:imagenet}, showing that the 
WCoRD+KD method achieves the best student performance on the ImageNet dataset.
The relative improvement of WCoRD over CRD on Top-1 error is $44.4\%$, and the relative improvement on Top-5 error is $23.08\%$, which further
demonstrates the scalability of our method.



\subsection{Cross-Modal Transfer}

\begin{table*}[t!]
\small
\begin{center}
\begin{tabular}{l|c|cc|cccc|cc}
\toprule
 & Parameter Size  & Teacher & Student & AT & SP & CC & CRD & LCKT & WCoRD \\
\midrule
ResNet-8x4 & 4.61 Mb  & - & 82.16 & 82.43 &  82.45 & 80.97 & 83.43 & 81.30 & \textbf{84.50}\\
\hline 
Inception-v3 & 84.61 Mb & - & 79.12 & 80.75 &  80.81 & 79.01 & 79.68 & 80.74 & \textbf{80.85}\\
\bottomrule
\end{tabular}
\caption{
\small{AUC (\%) of the ResNet-8x4 and Inception-v3 student networks on the OCT-GA dataset.}
}
\label{tb:privilege}
\end{center}
\vspace{-5mm}
\end{table*}

We consider a setting where one modality $\Xcal$ contains a large amount of labeled data, while the other modality $\Ycal$ has only a small labeled dataset. Transferring knowledge from $\Xcal$ to $\Ycal$ should help improve the performance of the model trained on $\Ycal$.
We use linear probing to evaluate the quality of the learned representation, a common practice proposed in~\cite{alain2016understanding, zhang2017split}. 
Following \cite{tian2019contrastive}, the same architecture is applied to both teacher and student networks. We first map images in the RGB space to the Lab color space (L: Luminance, ab: Chrominance), then train a ResNet18 (teacher) on the Luminance dimension of Labeled Tiny-ImageNet~\cite{deng2009imagenet}, which we call L-Net. The accuracy of L-Net on Tiny-ImageNet is 47.76$\%$. The student network, denoted as ab-Net, is trained on the Chrominance dimension of unlabeled STL-10 dataset~\cite{coates2011analysis}.

In experiments, we distill general knowledge from L-Net to ab-Net with different objective functions such as CRD and WCoRD.
Linear probing is performed by fixing the ab-Net. We then train a linear classification module on top of features extracted from different layers in the ResNet18 for 10-category classification. 
Results are summarized in Table~\ref{tb:crossmodal}. For reference, training student model from scratch with ResNet18 on STL-10 can reach an accuracy of $64.7\%$. 
From Table \ref{tb:crossmodal}, WCoRD outperforms CRD when using features extracted from the 2nd-4th residual blocks, indicating features extracted from these layers via WCoRD is more informative than those from CRD.

\subsection{Privileged Information Distillation}

In many real-world scenarios, large datasets required to train deep network models cannot be released publicly, due to privacy or security concerns.
One solution is privileged information distillation~\cite{vapnik2009new}: instead of releasing the original data, a model trained on the data is made public, which can serve as a teacher model.
Other researchers can train a student model on a smaller public dataset, leveraging this teacher model as additional supervision to enhance the performance of the student model.



For example, Geographic Atrophy (GA) is an advanced, vision-threatening form of age-related macular degeneration (AMD), currently affecting a significantly large number of individuals.
Optical coherence tomography (OCT) imaging is a popular method for diagnosing and treating many eye diseases, including GA~\cite{boyer2017pathophysiology}. 
To automatically detect GA in OCT images, a binary classifier can be trained with labeled data such as OCT-GA, an institutional dataset consisting of 44,520 optical coherence tomography (OCT) images of the retina of $1088$ patients; $9640$ of these images exhibit GA, and each image contains $512$ by $1000$ pixels. 


The resolution of images in OCT-GA is relatively low, and the small size of the dataset puts limitations on the learned model. One way to improve model performance is by leveraging additional larger high-resolution datasets~\cite{sun2017revisiting}.
Two challenges prevent us from doing this in real-life scenarios: ($i$) additional datasets may be private, and only a pre-trained model is publicly available; and ($ii$) the disease of interest may not be among those labeled categories in the additional datasets (\emph{e.g.}, GA may not be the focus of interest in other imaging datasets). 

One example is the OCT dataset introduced by \cite{kermany2018identifying}, consisting of 108,312 images from 4,686 patients for 4-way classification: choroidal neovascularization (CNV), diabetic macular edema (DME), Drusen, and Normal.
To test our proposed framework in privileged distillation setting, we treat this larger dataset as inaccessible and only use a pre-trained model as the teacher, as is the case in many real-life scenarios. Then we train a model on the smaller OCT-GA dataset as the student network, and use privileged distillation to transfer knowledge from the teacher to the student.

We test both Inception-v3 and ResNet8x4 models as student networks for GA disease identification. 
KL divergence cannot be used here, as both the learning goal and the training datasets for teacher and student networks are different.
This is designed to test the knowledge generalization ability of a model. 
As shown in Table \ref{tb:privilege}, WCoRD achieves an improvement of $2.34\%$ and $0.96\%$ compared to the basic student and CRD methods, respectively.
The relative improvement with ResNet8x4 is $\frac{\text{WCoRD} - \text{CRD}}{\text{CRD}-\text{AT}}=107.0\%$.
Since the goal of the teacher and student models are different, features from the teacher are biased.
When the student uses the same architecture as the teacher (Inception-v3), CRD performs worse than both AT and SPKD in Table \ref{tb:privilege}, which can be interpreted as low generalizability. With the help from LCKT, WCoRD is still able to obtain a comparable accuracy.
These results serve as strong evidence that WCoRD possesses better knowledge generalization ability than CRD. 
 
\begin{table}[t!]
\begin{center}
\resizebox{\columnwidth}{!}{
\begin{tabular}{l|cccccccc}
\toprule
 $\lambda_2$ & 0 & 0.05 & 0.06 & 0.07 & 0.08 & 0.09 & 0.1 & 0.2 \\
\midrule
Mean & 83.43 & 83.80 & 84.15    & \textbf{84.50}    & 83.83    & 83.70    & 83.65 & 82.28\\
Std  & 0.48 & 0.71  & 0.69    & 0.91    &  0.91   &  0.66   & 0.49    & 0.50\\

\bottomrule
\end{tabular}
}
\caption{
\small {AUC (\%) of student network ResNet-8x4 with different weights on the local knowledge transfer term.}
    }
\label{tb:privilege_ablation}
\end{center}
\vspace{-6mm}
\end{table} %

\vspace{5pt}
\noindent\textbf{Ablation Study}\,
We investigate the importance of the local knowledge transfer term  $\Lcal_{\text{LCKT}}(\cdot)$ in WCoRD. As shown in Tables \ref{tb:cifar100_same} and \ref{tb:cifar100_diff}, without it, WCoRD cannot consistently outperform CRD in different student-teacher architecture settings. 
By fixing $\lambda_1$ for the $\Lcal_{\text{GCKT}}(\cdot)$ loss with $\lambda_1=1$, we adjust the hyper-parameter $\lambda_2$ from $0.01$ to $0.2$. Table \ref{tb:privilege_ablation} reports the results, where it is evident that 
with $\lambda_2=0.07$ WCoRD performs the best. 
Also, we observe that when choosing $\lambda_2$ from $(0,0.1]$, it is consistently better than the model variant with $\lambda_2=0$. This indicates that our model is relatively robust given different hyper-parameter choices.

\section{Conclusions}

We present Wasserstein Contrastive Representation Distillation (WCoRD), a new framework for knowledge distillation.
WCoRD generalizes the concept of contrastive learning via the use of Wasserstein metric, and introduces an additional feature distribution matching term to further enhance the performance.
Experiments on a variety of tasks 
show that our new framework consistently improves the student model performance.
For future work, we plan to further extend our framework to other applications, such as federated learning \cite{konevcny2016federated} and adversarial robustness~\cite{papernot2016distillation}.

\newpage
{\small
    \bibliographystyle{ieee_fullname}
    \bibliography{ref}
}

\newpage
\twocolumn[{%
	\renewcommand\twocolumn[1][]{#1}%
	\appendix
	\begin{center}
		\centering
		\small
		\begin{tabular}{lccccccc}
			\toprule
			\thead{Teacher \\ Student} & \thead{WRN-40-2 \\ WRN-16-2} & \thead{WRN-40-2 \\ WRN-40-1} & \thead{resnet56\\resnet20} & \thead{resnet110\\resnet20} & \thead{resnet110\\resnet32} & \thead{resnet32x4\\resnet8x4} & \thead{vgg13\\vgg8}\\
			\midrule
			Teacher & 75.61 & 75.61 & 72.34 & 74.31 & 74.31 & 79.42 & 74.64 \\
			Student & 73.26 & 71.98 & 69.06 & 69.06 & 71.14 & 72.50 & 70.36 \\
			\midrule
			
			CRD & 75.48 $\pm$ 0.09& 74.14 $\pm$ 0.22& 71.16 $\pm$ 0.17& 71.46 $\pm$ 0.09& 73.48 $\pm$ 0.13& 75.51 $\pm$ 0.18& 73.94 $\pm$ 0.22\\
			
			CRD+KD & 75.64 $\pm$ 0.21& 74.38 $\pm$ 0.11& 71.63 $\pm$ 0.15& 71.56 $\pm$ 0.16& 73.75 $\pm$ 0.24& 75.46 $\pm$ 0.25& 74.29 $\pm$ 0.12\\
			
			WCoRD & 75.88$\pm$ 0.07 & {74.73} $\pm$ 0.17 & 71.56 $\pm$ 0.13 & 71.57 $\pm$ 0.09 & 73.81$\pm$ 0.11 & 75.95 $\pm$ 0.11 & 74.55$\pm$ 0.18 \\
			
			WCoRD+KD & {76.11} $\pm$ 0.15 & {74.72}$\pm$ 0.14 & {71.92}$\pm$ 0.17 & {71.88}$\pm$ 0.15 & {74.20}$\pm$ 0.20 & {76.15}$\pm$ 0.14 & {74.72}$\pm$ 0.13 \\
			
			\bottomrule
		\end{tabular}
		\captionof{table}{\small{Results with standard deviation of both the CRD and WCoRD methods.
				}}
		\label{tab:cifar100_with_std}
	\end{center}

	\begin{center}
		\centering
		\small
		\begin{tabular}{c|ccccccccccc}
			\toprule
			$\lambda_1$ & 0 & 0.05 & 0.07 & 0.1 & 0.15 & 0.2 & 0.5 & 0.8 & 1.0\\
			\midrule
			Result & 79.12 & 80.11 & 82.15 &  {83.50} & 83.33 & 83.78 & 84.2 & \textbf{84.5} & 84.3\\
			\bottomrule
		\end{tabular}
		\captionof{table}{\small{AUC (\%) of student network ResNet-8x4 with different $\lambda_1$ values on the GCKT term.
		}}
		\label{tab:ablation_supp}
	\end{center}
}]

\section{WCoRD Algorithm}
The detailed implementation of the proposed Wasserstein Contrastive Representation Distillation (WCoRD) method is summarized in Algorithm \ref{alg:wf}.
\begin{algorithm}[!h]
\caption{The proposed WCoRD Algorithm.}
\label{alg:wf}
\begin{algorithmic}[1]
\STATE {\bfseries Input:} A mini-batch of data samples $\{\xv_i, y_i\}_{i=1}^n$.
\STATE Extract features $\hv^T$ and $\hv^S$ from the teacher and student networks, respectively.
\STATE Construct a memory buffer $\Bcal$ to store previous computed features.
\STATE Global contrastive knowledge transfer: 
\STATE \quad Max. the GCKT loss in Eqn. (11) over $\thetav_S$ and $\phiv$.
\STATE Local contrastive knowledge transfer: 
\STATE \quad Min. the LCKT loss in Eqn. (13) over $\thetav_S$.
\STATE Min. the task-specific loss over $\thetav_S$.
\end{algorithmic}
\end{algorithm}

\section{Baseline Methods and Model Architectures}
\subsection{Baseline Methods}
We compare WCoRD with a number of baseline distillation methods, detailed below.
\begin{itemize}
	\item Fitnets: Hints for thin deep nets \cite{romero2014fitnets};
	\item Knowledge Distillation (KD) \cite{hinton2015distilling};
	\item Attention Transfer (AT) \cite{zagoruyko2016paying};
	\item Like what you like: Knowledge distilllation via neuron selectivity transfer (NST) \cite{huang2017like};
	\item A gift from knowledge distillation: fast optimization, network minimization and transfer learning (FSP) \cite{yim2017gift};
	\item Learning deep representations with probabilistic knowledge transfer (PKT) \cite{passalis2018learning};
	\item Paraphrasing complex network: network compression via factor transfer (FT) \cite{kim2018paraphrasing};
	\item Similarity-preserving knowledge distillation (SP) \cite{tung2019similarity};
	\item Correlation congruence (CC) \cite{peng2019correlation};
	\item Variational information distillation for knowledge transfer (VID) \cite{ahn2019variational};
	\item Relational knowledge distillation (RKD) \cite{park2019relational};
	\item Knowledge transfer via distillation of activation boundaries formed by hidden neurons (AB) \cite{heo2019knowledge};
	\item Contrastive representation distillation (CRD) \cite{tian2019contrastive} via NCE~\cite{gutmann2010noise}.
\end{itemize}
Note that the hyper-parameter setup for these baseline methods follows the setup in CRD~\cite{tian2019contrastive}.
\subsection{Model Architectures}
In experiments, we utilize the following model architectures.
\begin{itemize}
	\item Wide Residual Network (WRN)~\cite{zagoruyko2016wide}: WRN-$d$-$w$ represents wide ResNet with depth $d$ and width factor $w$.
	\item resnet~\cite{he2016deep}: We use ResNet-$d$ to represent CIFAR-style resnet with 3 groups of basic blocks, each with $16$, $32$, and $64$ channels, respectively. In our experiments, resnet8x4 and resnet32x4 indicate a 4 times wider network (namely, with $64$, $128$, and $256$ channels for each of the blocks).
	\item ResNet~\cite{he2016deep}: ResNet-$d$ represents ImageNet-style ResNet with bottleneck blocks and more channels. 
	\item MobileNetV2~\cite{sandler2018mobilenetv2}: In our experiments, we use a width multiplier of $0.5$.
	\item vgg~\cite{simonyan2014very}: The vgg networks used in our experiments are adapted from their original ImageNet counterpart.
	\item ShuffleNetV1~\cite{zhang2018shufflenet}, ShuffleNetV2~\cite{tan2019mnasnet}: ShuffleNets are proposed for efficient training and we adapt them to input of size 32x32.
	\item InceptionNet-v3 \cite{szegedy2016rethinking} is used for the teacher network in the priviledge distillation experiment.
\end{itemize}

\section{Additional Results}
In Table~\ref{tab:cifar100_with_std}, we report additional results of the baseline distillation methods when combined with KD, and the standard deviation of the results of both CRD and WCoRD, with or without KD. Our method achieves better performance.

We also tested the importance of the GCKT module in WCoRD. We fixed the LCKT module by choosing $\lambda_2=0.1$, and then we adjust $\lambda_1$ from $0$ to $1.0$. Results are summarized in Table~\ref{tab:ablation_supp}.
Our model is fairly robust towards different choices of $\lambda_1$. Also, without the help of the GCKT module, models only with LCKT cannot obtain a very good performance.

\end{document}